# PHM-Bench: A Domain-Specific Benchmarking Framework for Systematic Evaluation of Large Models in Prognostics and Health Management

1st Puyu Yang[1,2,3,4], 2nd Laifa Tao[1,2,3,4], 3rd Zijian Huang[2,3,4], 4th Haifei Liu [2,3,4], 5th Wenyan Cao [2,3,4], 6th Hao Ji[1], 7th Jianan Qiu[1], 8th Qixuan Huang [2,3,4], 9th Xuanyuan Su[1], 10th Yuhang Xie[2,3,4], 11th Jun Zhang[2,3,4], 12th Shangyu Li[1,2,3,4], 13th Chen Lu[1,2,3,4] , 14th Zhixuan Lian[1,*]

[1]Hangzhou International Innovation Institute, Beihang University, China
[2]Institute of Reliability Engineering, Beihang University, Beijing, China
[3]Science & Technology on Reliability & Environmental Engineering Laboratory, Beijing, China
[4]School of Reliability and Systems Engineering, Beihang University, Beijing, China
*Corresponding author, lianzhixuan@buaa.edu.cn[1]

Puyu Yang: 20212202@cqu.edu.cn; 0009-0007-5445-0209
Laifa Tao: taolaifa@buaa.edu.cn.; 0000-0002-9232-7216
Zijian Huang; hzj1234@buaa.edu.cn; 0009-0007-5442-7017
Haifei Liu: phoebeliu@buaa.edu.cn; 0000-0002-3099-7185
Wenyan Cao: 20276030@buaa.edu.cn; 0009-0000-2883-58926
Hao Ji: zy2457529@buaa.edu.cn:0009-0002-1099-4515
Jianan Qiu: ZY2457230@buaa.edu.cn; 0009-0009-5784-200X8
Qixuan Huang: qxhuang@buaa.edu.cn; 0000-0002-5376-9927
Xuanyuan Su: suxuanyuan@buaa.edu.cn; 0000-0003-0306-183X10
Yuhang Xie: 1953179220@qq.com; 0009-0000-9367-699811
Jun Zhang: znj0089@163.com; 0009-0004-4539-085512
Shangyu Li: lishangyu@buaa.edu.cn; 0000-0003-1426-9206
Chen Lu: luchen@buaa.edu.cn
Zhixuan Lian: lianzhixuan@buaa.edu.cn; 0000-0003-0878-9344

# Abstract

With the rapid advancement of generative artificial intelligence, large language models (LLMs) are increasingly adopted in industrial domains, offering new opportunities for Prognostics and Health Management (PHM), addressing challenges such as high development costs, long deployment cycles, and limited generalizability. However, despite the growing synergy between PHM and LLMs, existing evaluation methodologies often fall short regarding structural completeness, dimensional comprehensiveness, and evaluation granularity, severely hampering the in-depth integration of LLMs into the PHM domain.

To address these limitations, this study, drawing upon two decades of PHM research and recent advancements in LLM-driven PHM systems, proposes PHM-Bench, a novel three-dimensional evaluation framework for PHM-oriented large models. Grounded in the triadic structure of fundamental capability, core task, and entire lifecycle, PHM-Bench is designed specifically for the unique demands of PHM system engineering. It systematically defines multi-level evaluation metrics spanning knowledge comprehension, algorithmic generation, task optimization, etc., aligning with typical PHM tasks including condition monitoring, fault diagnosis, fault & RUL prediction, and maintenance decision-making, thus establishing a comprehensive assessment mechanism, bridging complex engineering systems' design, development, and operational stages.

Utilizing both self-constructed case sets and publicly available industrial datasets, PHM-Bench enables multi-dimensional evaluation of general-purpose and domain-specific models across diverse PHM tasks. Experiments demonstrate its effectiveness in revealing model capabilities and limitations, distinguishing performance across tasks, and providing a unified baseline for model development and optimization. PHM-Bench lays the methodological foundation for industrial-scale assessment of LLMs in PHM and offers a critical benchmark to guide the transition from general-purpose to PHM-specialized models.

# Keywords

Prognostics and Health Management (PHM); Large Language Model (LLM); Domain-Specific Evaluation; Benchmarking Framework; Multi-Dimensional Evaluation Metrics

# 1. Introduction

Prognostics and Health Management (PHM) integrates sensor measurements, data-science techniques, and artificial intelligence with expert knowledge, failure-physics models, and maintenance-support information to monitor, diagnose, predict, evaluate, and manage equipment health, forming the foundation for reliable operation, efficient production, high availability, and mission safety of complex assets (Y.-F. Li et al., 2024; Tao, Li, et al., 2025; Zio, 2022). The rapid advance of equipment intelligence, accelerated model-iteration cycles, increasingly complex operating conditions, and diverse data modalities now demand reduced expert dependence, shorter development timelines, enhanced generalization, and multimodal fusion. Traditional PHM systems, however, are hampered by lengthy development, slow responsiveness, high cost, and limited generalizability, which constrain performance and real-world efficacy(El-Brawany et al., 2023). With the advent of general AI, generative large models—endowed with powerful generalization and synthesis capabilities in semantic understanding, code generation, logical reasoning, and multimodal processing—promise to address these bottlenecks, enabling minimal expert intervention, efficient development, cross-scenario adaptability, and cohesive multimodal collaboration.

Recent prototypes and preliminary products have begun integrating large models with PHM technologies, such as the Industrial Metaverse for Smart Manufacturing (Ren et al., 2024), Argonne National Laboratory's LLM-based complex-system explainable fault diagnosis (Dave et al., 2024), Baidu Qianfan's Equipment Health Management and Predictive Maintenance Platform, and the ChatGLM2-6B-chat-powered, FaultsMind bearing-fault diagnosis model (Tao, Liu, et al., 2025). These initiatives leverage LLMs' language understanding and generation capabilities alongside domain knowledge and data to achieve initial improvements in fault diagnosis, remaining-useful-life prediction, and maintenance decision-making, thereby demonstrating generative, generalization, and adaptive-learning potential in specific settings.

However, despite ongoing advances in PHM–LLM research and applications, evaluation and validation approaches remain conventional and fragmented. From the entire lifecycle viewpoint, most existing assessments focus on late-stage operational performance—applying general PHM metrics to task outcomes (C. Chen et al., 2021; Deng et al., 2023; Jose et al., 2024; Le Xuan et al., 2024; H.-W. Lu & Lee, 2022a; Pinciroli et al., 2022; B. Yan et al., 2021a; Zhao et al., 2024)—or design bespoke measures for individual models (D. Chen et al., 2024; Ding et al., 2021; Diversi & Speciale, 2024; Jia et al., 2016; Kamariotis et al., 2024; Koutroulis et al., 2022;

Moradi et al., 2023; Soualhi et al., 2024; X. Yang et al., 2025; Yu et al., 2023; Y. Zhang et al., 2024; J. Zhou et al., 2024). Such siloed evaluation frameworks lack the comprehensiveness, dimensionality, and granularity required to assess end-to-end PHM applications across business scenarios, operating conditions, target characteristics, and data modalities, thereby impeding quantification of LLMs' value, holistic performance benchmarking, and iterative optimization of PHM systems.

To address these challenges and achieve an integrated, comprehensive, and finely grained evaluation framework—thereby fostering the scientific, robust, and rapid advancement of PHM-specific large models—this paper, drawing on our team's nearly twenty years of PHM research and recent experience in health-management large-model development, proposes PHM-Bench: a task-scenario-driven evaluation framework tailored to PHM system engineering requirements and the unique characteristics of both large models and the PHM domain. Centered on a three-dimensional "Foundational Capability–Core Task–Entire Lifecycle" structure, PHM-Bench targets key PHM tasks (condition monitoring, fault diagnosis, fault&RUL prediction and maintenance decision-making) and, across the sub-dimensions of knowledge comprehension, retrieval and generation, algorithm generation and recommendation, and task plan generation and optimization, constructs a multi-tiered metric system that spans the design, development and deployment stages—providing explicit guidance for both large-model development and industrial application.

Building on this, we leverage a proprietary case collection and open industrial datasets to conduct systematic, multi-dimensional evaluations of several state-of-the-art general models alongside domain-specific industrial models. Our results demonstrate that PHM-Bench effectively exposes each model's strengths and limitations in varied task scenarios, illuminates performance differentials in representative PHM business contexts, and establishes unified standards and baselines for PHM large-model design, optimization, and deployment. As such, PHM-Bench lays the methodological groundwork for industrial-scale PHM system evaluation and serves as a critical benchmark for the evolution from general-purpose large models to PHM-focused specialized models.

In this work, we make the following key contributions:

1. We propose a pioneering evaluation framework and metric system—PHM-Bench—specifically designed for assessing large models in the Prognostics and Health Management (PHM) domain.

2. We construct optimized benchmark datasets and streamlined evaluation pipelines tailored to PHM tasks.

3. We establish domain-specific performance baselines for PHM-related tasks and conduct comprehensive performance evaluations.

Through the development of the PHM-Bench framework, this study aims to provide both theoretical foundations and practical guidance for advancing AI-enabled PHM systems. We advocate a paradigm shift from traditional PHM approaches—characterized by "manual design, post-hoc validation, and localized feedback"—to an intelligent co-engineering paradigm centered on "AI-assisted design, integrated evaluation, and global optimization feedback." This transformation supports the design, development, validation, and deployment of PHM systems under the emerging foundation model architecture. The resulting methodology introduces new tools, technologies, and platforms for PHM research, enabling a generational leap from customized to generalized solutions, from discriminative to generative modeling, and from idealized experimentation to real-world application.

# 2 Related Work

## 2.1 Entire Lifecycle PHM

PHM systems underpin entire lifecycle equipment health management, yet their traditional design and development rely on multi-stage manual analysis and model construction, encountering bottlenecks such as expert dependency, knowledge fragmentation, data paucity, and validation difficulties (H. Li, Cao, et al., 2024; Zheng et al., 2025; K. Zhou, Lu, et al., 2025). Large language models (LLMs) promise to automate cumbersome, low-efficiency tasks—solution evaluation, functional verification, algorithm development, and testing, thereby improving PHM design efficiency and in-service performance (Tao, Li, et al., 2025). Under a PHM–LLM collaborative paradigm, LLM deployment should span all PHM phases—design, development, and service—so their evaluation must likewise cover the entire lifecycle. Since current PHM validation emphasizes late-stage task performance and in-service assessment, we propose an entire lifecycle large-model evaluation framework that illustrates how LLM integration across all stages can fundamentally optimize PHM design and validation, providing a holistic perspective for future research, application benchmarking, optimization, and novel PHM architectures.

## 2.2 PHM Core Task Evaluation

As a paradigmatic industrial enabling technology, PHM requires evaluation methods that align closely with real-world operational demands Its typical tasks have been studied extensively with task-specific metrics. For fault diagnosis, confusion matrix–based accuracy and imbalance-sensitive indicators such as precision, recall, F1-score, and Cohen's kappa are widely used(Su et al., 2024; W. Zhang et al., 2021); for RUL prediction, mean absolute error (MAE) and root-mean-square error (RMSE) are predominant(B. Yan et al., 2021b; L. Yang et al., 2024a), etc. While these localized measures provide task-level insights, they remain fragmented and lack coherence within a unified assessment paradigm. To address this gap, we introduce a comprehensive evaluation framework for large models, grounded in PHM's core tasks and cross-task requirements, thereby supporting systematic and holistic benchmarking.

## 2.3 Large Models Fundamental Capability

Recent Transformer-based large models have achieved significant advances on general foundational benchmarks (Achiam et al., 2023). Standardized evaluations now exist for knowledge storage and retrieval (Imtiaz Ahmed, 2025) language comprehension and generation (X. Liu et al., 2024), logical reasoning (Chen Z;Luo, 2023), and cross-modal integration (Merkelbach et al., n.d.; Wu JL;Lu, 2022). However, these general metrics diverge markedly from PHM's engineering requirements—namely, domain-intensive knowledge, sensitivity to temporal signals, and decision-making reliability (Kamariotis A;Tatsis, 2024)—and thus cannot effectively evaluate, compare, optimize, or validate large models in PHM's vertical applications. Accordingly, we propose a PHM-oriented foundational capability assessment framework for large models to enable fine-grained, scenario-specific evaluation and to ensure the scientific rigor and relevance of PHM model benchmarking.

### 2.4 Gap Analysis

Existing PHM validation efforts have largely targeted end-stage task performance, with insufficient attention to the design, development, and maturation phases. This gap stems partly from underestimating early design's significance and partly from the reliance on expert judgment and qualitative analysis in front- and mid-stage workflows, which impedes the creation of actionable, quantitative evaluation schemes. The advent of generative large models (LLMs) now enables a closed-loop, full-lifecycle evaluation—from design through validation to

optimization—a key innovation of this work. Accordingly, we propose PHM-Bench, the first domain-driven, theoretically grounded, and practice-oriented framework for comprehensive large-model assessment in PHM. From a tri-axis "Foundational Capability–Core Task–Entire Lifecycle" perspective, PHM-Bench integrates task requirements, model attributes, and engineering realities into a clear, quantifiable, and extensible evaluation system, underpinned by standardized datasets, automated workflows, and expert review. Empirical studies confirm its effectiveness in model benchmarking, capability diagnosis, and optimization guidance.

# 3 PHM-Bench Evaluation Framework

This section develops a comprehensive evaluation framework for large-model PHM capabilities by first establishing its design principles and architectural blueprint (Section 3.1), then defining the evaluation criteria and dimensional taxonomy (Section 3.2), and finally detailing the assessment methods and procedural workflows for end-to-end performance analysis (Section 3.3). By unifying framework construction, metric specification, and methodological execution into a coherent logical chain, it ensures operational feasibility and real-world applicability and provides a rigorous foundation for systematic, interpretable cross-model comparisons, thereby facilitating the development of robust and efficient industrial PHM solutions.

## 3.1 PHM-Bench framework construction

This section presents the design of the PHM-Bench evaluation framework, which delivers a scientific, systematic, and comprehensive methodology for assessing large-model performance in PHM, addressing challenges such as complex operating conditions, multimodal data, and task-specific requirements. As illustrated in Fig. 1, PHM-Bench adopts a four-layer, modular architecture comprising the Input Layer, Model Layer, Evaluation Layer, and Capability Engine.

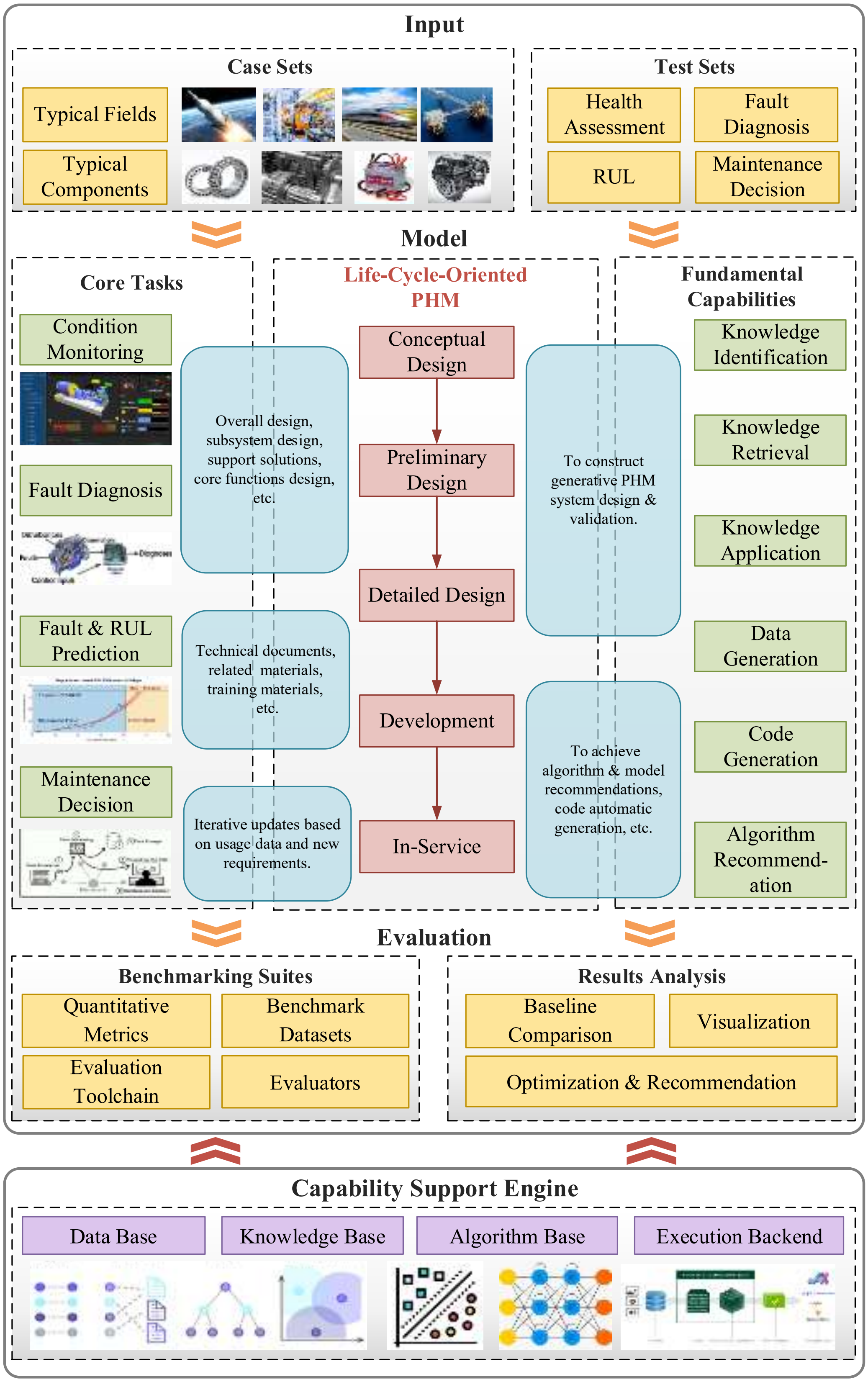

**Fig.1** Overall Architecture of PHM-Bench

**(1) Input Layer：**

The input layer establishes the foundational datasets and test tasks for large-model PHM evaluation. It ingests all required data and configurations—test cases paired with their corresponding tasks—and forwards them to the model layer. By systematically surveying high-impact PHM publications from the past three years, we distilled operational contexts, objectives, data characteristics and fault modalities to create a standardized case template and assemble a comprehensive test suite spanning key industries (aerospace, maritime, defense) and 18 representative components/subsystems (e.g., batteries, motors, gears, bearings). From this repository, task-specific test sets were derived to cover fault diagnosis, degradation forecasting, health monitoring, and maintenance decision-making. The input layer then standardizes these inputs and, when necessary, decomposes complex evaluations into manageable sub-tasks, thereby ensuring consistency and reliability across the entire assessment process.

**(2) Model layer：**

The model layer—PHM-Bench's core—orchestrates task execution and capability representation across the PHM system lifecycle. Centered on five lifecycle phases (conceptual design, preliminary design, detailed design, development, and service), it aligns model evaluation with each stage's tasks and requisite capabilities. The Core Tasks module addresses key PHM functions—condition monitoring, fault diagnosis, fault&RUL prediction, and maintenance decision-making—reflecting essential design and development requirements. The Fundamental Capabilities module specifies the large model's supporting proficiencies—domain knowledge acquisition, comprehension, retrieval, and application; data and code generation for real-world scenarios; and optimal algorithm recommendation—across analysis, design, verification, and deployment activities. By interweaving task flows with capability streams along the entire lifecycle, the model layer establishes a coherent evaluation logic and granular testing tasks, thereby enabling PHM-Bench to deliver comprehensive multidimensional performance assessments.

**(3) Evaluation layer：**

The evaluation layer establishes a standardized assessment framework grounded in large-model PHM capability metrics and comprising automated and expert evaluations, as well as comprehensive result analysis and feedback, to guarantee objectivity, completeness, and interpretability. The automated module employs the three-dimensional PHM taxonomy—foundational capabilities, core tasks, and entire lifecycle performance—alongside trillion-parameter models within an automated toolchain to quantitatively measure PHM capabilities. For metrics that resist full automation or require post-automation validation, a human evaluation module enlists domain experts to confirm interpretability and practical applicability. Finally, the result analysis component provides in-depth analytics and visualizations, mapping evaluation objectives to functional modules and delivering comparative performance metrics, task-level outcomes, and targeted optimization recommendations, thereby ensuring robust, reliable assessments and enhancing the deployment of large models in PHM.

**(4) Engine layer:**

The Capability Support Engine underpins the PHM-Bench evaluation framework by integrating standardized industrial datasets, a structured PHM knowledge base, an algorithm library of both conventional and advanced PHM models, and a comprehensive testing environment for deployment, resource scheduling, distributed execution and logging, thereby ensuring the scientific rigor, completeness and operational reliability of the entire evaluation process.

## 3.2 Dimensional Structure and Metric System of Evaluation

Given PHM's engineering-oriented, industrial focus, we introduce a three-dimensional evaluation framework for large-scale PHM models—"Capability Base," "Task Efficiency" and "System Collaboration"—corresponding to the core-task (X), foundational-capability (Y) and entire-lifecycle (Z) axes, and refine a detailed metrics taxonomy accordingly (Fig. 2).

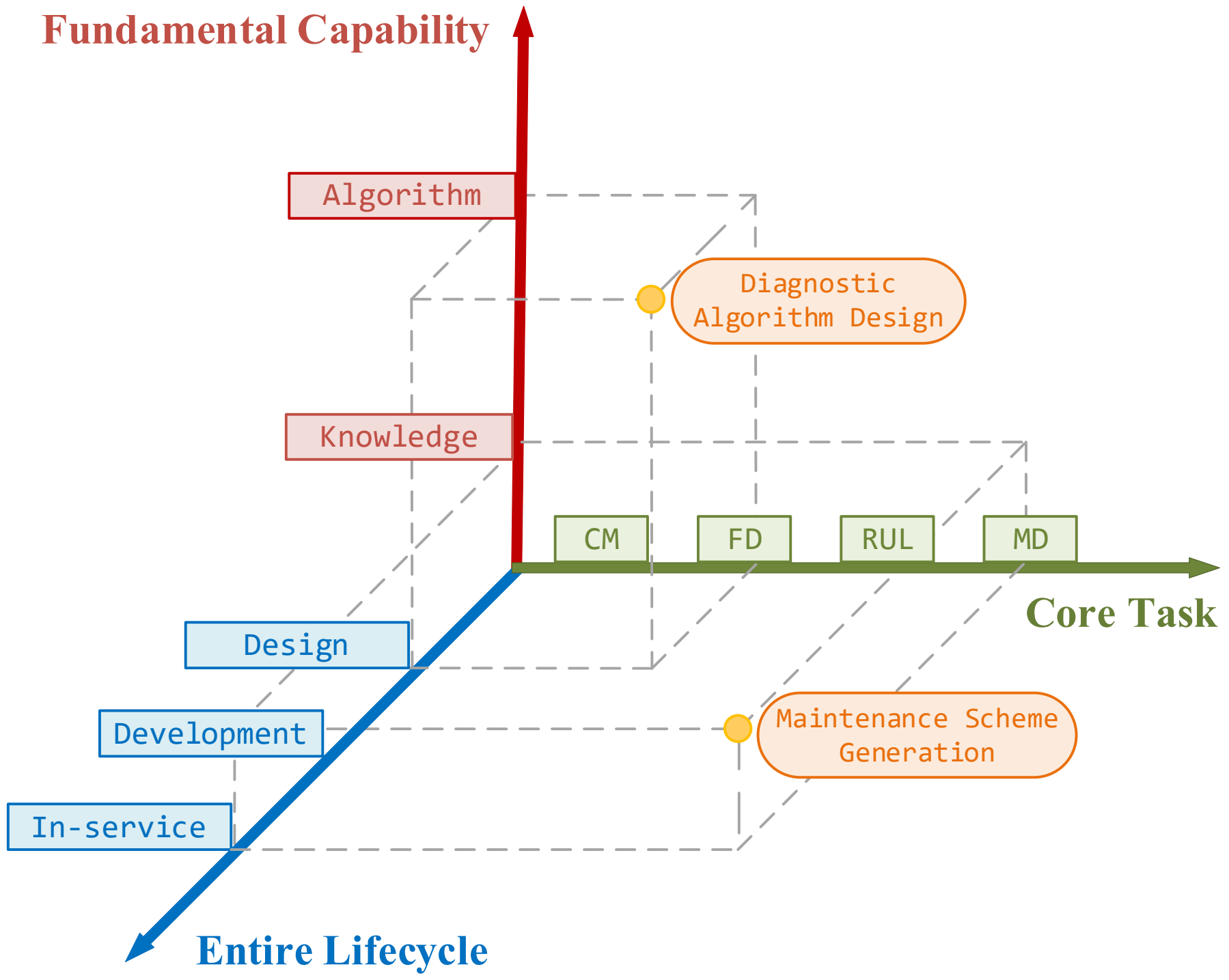

**Fig. 2** Dimensional Breakdown of PHM-Bench Evaluation

We then explore the capability requirements for each dimension, presenting for each the specific indicators and their definitions.

### 3.2.1 Core Task Dimension

In Section 2, we noted that PHM's task evaluation must align with real-world operational characteristics, yet existing assessments remain fragmented and task-specific. Here, we first summarize the common metrics for the four core tasks—condition monitoring, fault diagnosis, remaining-useful-life prediction, and maintenance decision-making, as demonstrated in Fig.3—and then propose a unified large-model evaluation framework that addresses four key business requirements of complex PHM solutions to ensure a comprehensive, systematic appraisal.

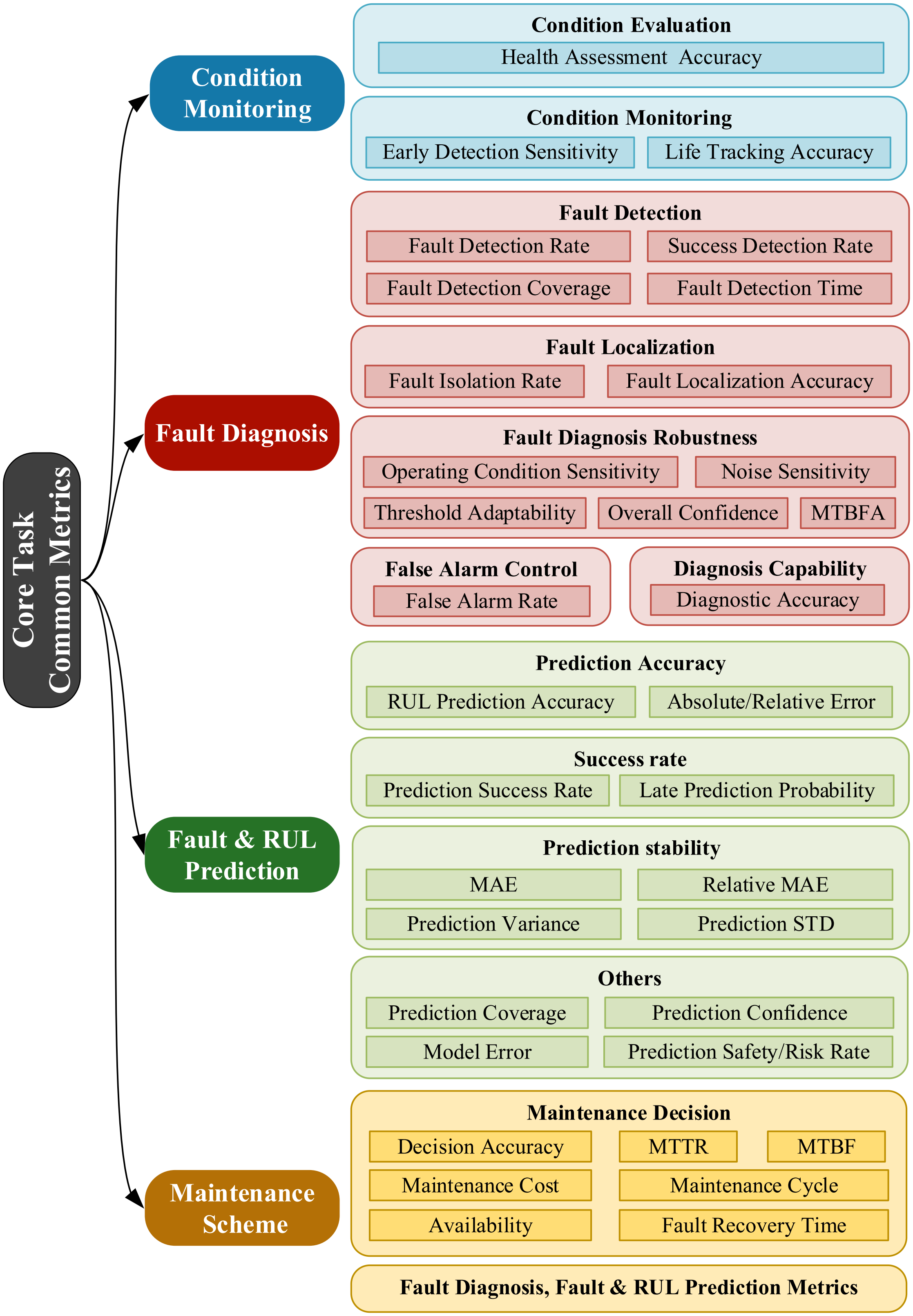

**Fig. 3** Summary of Core Task Common Metrics of PHM

Fig.3 presents a comprehensive framework that consolidates commonly used task capacity validation metrics of PHM, drawn from GJB military standards, ISO international standards, SAE-J aerospace standards, and NASA's industrial assessment schemes. This ensures both engineering practicality and alignment with industry benchmarks.

Tasks are categorized according to established PHM technical conventions, segmented into three primary functions: Condition Monitoring, Fault Diagnosis, Fault & RUL prediction, and Maintenance Decision(Ding et al., 2024; Kim et al., 2025; Mao et al., 2022; Xu et al., 2023). These categories are color-coded in the figure for clarity.

At the indicator level, the framework includes over 40 representative engineering metrics such as fault detection rate, fault isolation rate, MTBFA, RUL prediction accuracy, prediction stability, decision-making accuracy, and lifecycle tracking capability. These indicators are typically measurable and reproducible in real PHM systems via scenario simulation or embedded system implementation.

Unlike traditional system verification approaches focused on module-level functionality, this framework emphasizes a structure driven by task relevance, stage-specific adaptation, and indicator support. It evaluates the engineering adaptability of LLM-generated solutions and their ability to indirectly fulfill PHM tasks, serving as a basis for interpretability and capability transfer assessments.

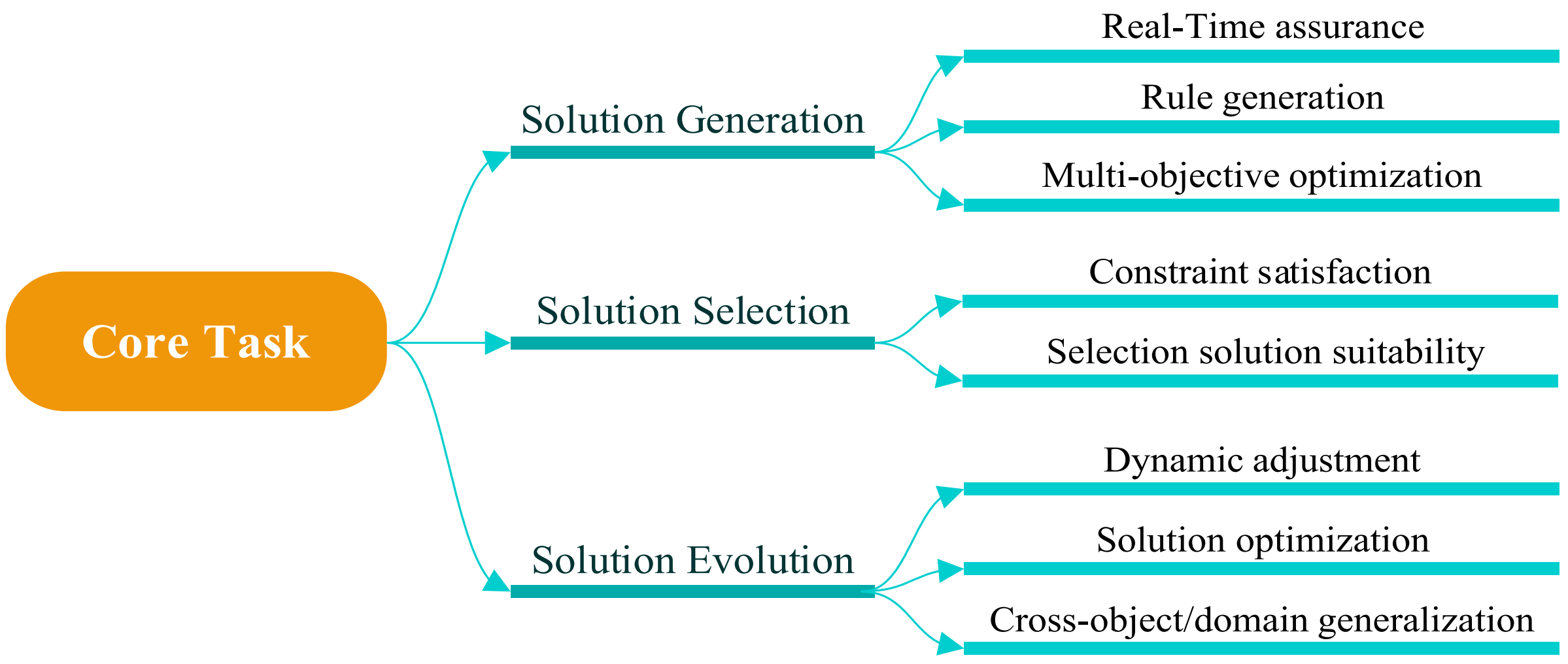

**Fig. 4** Evaluation Metric System for Core Task Dimension

To address the common challenges of complex PHM tasks, we focus on four core functions: Condition monitoring(Compare et al., 2022; Mei et al., 2022), Fault diagnosis(G. Liu & Wu, 2024; Souza et al., 2021), Fault & RUL prediction(Wen et al., 2025; J. Yan et al., 2022), and Maintenance Decision-making(Dong et al., 2025; C. Huang et al., 2024). We assess Large Model capabilities across three dimensions—Solution Generation, Solution Selection, and Solution Evolution—tailored to PHM-specific needs, as Fig.4.

**(1) Solution Generation**

Generating viable solutions for complex tasks is a fundamental capability of PHM-oriented LLMs (Tao et al., 2024). Effective evaluation must consider the specific task context, including deployment conditions, operational scenarios, and object characteristics. The model should autonomously generate customized PHM solutions that are feasible, targeted, and adaptable, accounting for complex dependencies (Xia et al., 2024). We propose the following indicators to quantify this capability:

**1) Task Adaptability Score (TAS）**

$$TAS = \frac{N_{\text{covered}}}{N_{\text{required}}}$$

In which:

$N_{\text{required}}$ : The number of target elements defined in the preset task description, such as goals, data sources, model types, deployment constraints, etc;

$N_{\text{covered}}$ : The number of target elements successfully reflected in the model output.

PHM tasks often involve complex structures, dynamic conditions, and multi-objective coordination. The model should automatically generate comprehensive PHM solution frameworks—including deployment configurations, algorithm architecture, and key performance metrics (Tao, Liu, et al., 2025). The Task Adaptability Score (TAS) evaluates the extent to which the model’s output aligns with all elements of the task requirements, beyond code generation or algorithm recommendation.

**2) Diagnostic Rule Generation Accuracy (DRGA)**

$$DRGA = \frac{N_{\text{valid_rules}}}{N_{\text{total_rules}}}$$

In which:

$N_{\text{total_rules}}$: The total number of diagnostic rules generated by the model;

$N_{\text{valid_rules}}$: The number of effective rules that can successfully identify faults in the validation set.

PHM tasks frequently rely on rule-based anomaly detection, causal inference, and logical reasoning (Suo et al., 2020). The Diagnostic Rule Generation Accuracy (DRGA) measures the model's ability to produce valid, executable fault diagnosis rules.

**3) Engineering Constraint Satisfaction Rate (ECS)**

$$ECS = \frac{1}{n}\sum_{i=1}^{n}\mathbb{I}\left(C_i\right)$$

In which:

$n$: The number of constraint conditions preset for the task;

$C_i$: The $i$-th constraint terms;

$\mathbb{I}\left(C_i\right)$: 1 if the model output satisfies this constraint, 0 otherwise.

Real-world PHM solutions must meet various engineering constraints such as real-time performance, memory limits, and interface standards (Polenghi et al., 2022). The Engineering Constraint Score (ECS) evaluates whether the model-generated solutions address deployment requirements logically and structurally.

**(2) Solution Selection**

When pre-existing solutions, industry strategies, or engineering precedents are available, the LLM must discern task constraints and context to identify and recommend the most appropriate option. This requires not only knowledge retrieval but also nuanced interpretation of constraints. To assess model performance in complex task-driven solution selection, we introduce the following indicators:

**1) Solution Selection Compatibility (SSC)**

$$SSC = \frac{1}{\left|S_{rel}\right|}\sum_{s_j \in S_{rel}}\eta\left(s_j\right)$$

In which:

$\mathcal{S}_{\mathrm{rel}}$: Set of optional solutions semantically related to the current task (provided by manual annotation or a benchmark system);

$\eta\left(s_j\right)$: Adaptation determination function, which means if the model successfully selects the $i$-th relevant solution and this solution can meet the task requirements, the result is 1; otherwise, it is 0.

$|\mathcal{S}_{\mathrm{rel}}|$: Total number of effective relevant solutions.

PHM systems often face multiple existing options, such as fault diagnosis algorithms (Zou et al., 2023) or maintenance strategies (Tao et al., 2024). The Solution Selection Compatibility (SSC) measures the model's ability to recommend the most semantically and logically suitable option based on task constraints and scenario features.

**2) Multi-Objective Balance Ratio (MOBR)**

$$MOBR = \frac{1}{k}\sum_{i=1}^{k}\frac{M_i}{M_i^{\max}}$$

In which:

$k$: Total number of objectives (usually 2 or 3, such as performance, computing time, memory footprint, etc.);

$M_i$: The performance of the current recommended solution on the $i$-th objective;

$M_i^{\max}$: The optimal value of the $i$-th target under the current task among all candidate solutions;

$MOBR \in [0,1]$: The closer it is to 1, the closer the solution is to the Pareto frontier (Momma et al., 2022).

In practice, PHM systems must balance multiple objectives, such as predictive accuracy, resource usage, and deployment speed (Desmet C;Cook, 2024). The Multi-Objective Balance Ratio (MOBR) assesses whether the model selects a solution that effectively balances these competing goals rather than maximizing one at the expense of others.

**(3) Solution Optimization**

PHM systems must adapt to evolving operating conditions and shifting task requirements. LLMs should support continuous plan adjustment (Shen et al., 2025), iterative refinement, and generalization (Polenghi et al., 2022), including performance retention in previously unseen scenarios (cold start), thus supporting the adaptive optimization of task solutions throughout the lifecycle of the PHM system. The following metrics evaluate these optimization capabilities:

**1) Priority Adjustment Success Rate (PASR)**

$$PASR = \frac{1}{N} \sum_{i=1}^{N} \mathbb{I}\left(\gamma_i \wedge \theta_i \wedge \omega_i\right)$$

In which:

$N$ : There is a need to adjust the priority of the task;

$\gamma_i$ : There is a need to adjust the priority of the task;

$\theta_i$ : The model has completed the priority adjustment;

$\omega_i$ : The adjusted task was executed successfully on schedule;

$\mathbb{I}(\cdot)$ : If the three conditions are all met, record 1; otherwise, record 0.

PHM often involves dynamic scheduling and task coordination (H.-W. Lu & Lee, 2022b), such as reprioritizing maintenance after health deterioration or rescheduling across systems. The Priority Adjustment Success Rate (PASR) gauges whether the model correctly identifies priority shifts and successfully implements adjustments.

**2) Optimization Gain Rate (OGR)**

$$OGR = \frac{P_{\text{after}} - P_{\text{before}}}{P_{\text{before}}}$$

In which:

$P_{\text{before}}$ : Performance indicators before optimization (such as accuracy, F1, RMSE, etc.);

$P_{\text{after}}$ : Optimized similar performance indicators.

A key goal of optimization is improving PHM model performance, such as diagnostic accuracy or execution speed. The Optimization Gain Rate (OGR) quantifies improvement in core metrics before and after optimization, reflecting the effectiveness of model-led enhancements.

**3) Optimization Cycle Efficiency (OCE)**

$$OCE = \frac{P_{\text{final}} - P_{\text{init}}}{n}$$

In which:

$P_{\text{final}}$ : The performance of the final plan;

$P_{\text{init}}$ : Initial scheme performance;

$n$ : Optimize the number of rounds (including the number of updates between the initial and final states).

Optimization in engineering must consider both performance and cost. The Optimization Cycle Efficiency (OCE) assesses the average performance gain per iteration, indicating the cost-effectiveness of the model's strategy.

**4) Cross-Domain Generalization Index (CDGI）**

$$CDGI = \frac{P_{\text{target}}}{P_{\text{source}}}$$

In which:

$P_{\text{source}}$ : Train or familiarize with performance metrics in the scenario;

$P_{\text{source}}$ : The closer it is to 1, the stronger the generalization ability;

$CDGI \in [0,1+]$ : The closer it is to 1, the stronger the generalization ability.

PHM strategies often lack generalizability. With LLM support, PHM solutions should adapt across equipment or scenarios. The Cross-Domain Generalization Index (CDGI) evaluates whether the model can output valid, adaptable solutions when faced with novel situations (Y. Cao et al., 2024a).

### 3.2.2 Foundational Capability Dimension

A model's foundational capabilities determine its effectiveness in supporting PHM applications. This section presents a domain-specific evaluation system assessing LLMs' foundational abilities in the PHM context. The framework comprises two sub-dimensions: Knowledge and Algorithm. It supports fine-grained evaluation of an LLM's ability to bridge theory and practice in PHM applications, as Fig.5 demonstrates.

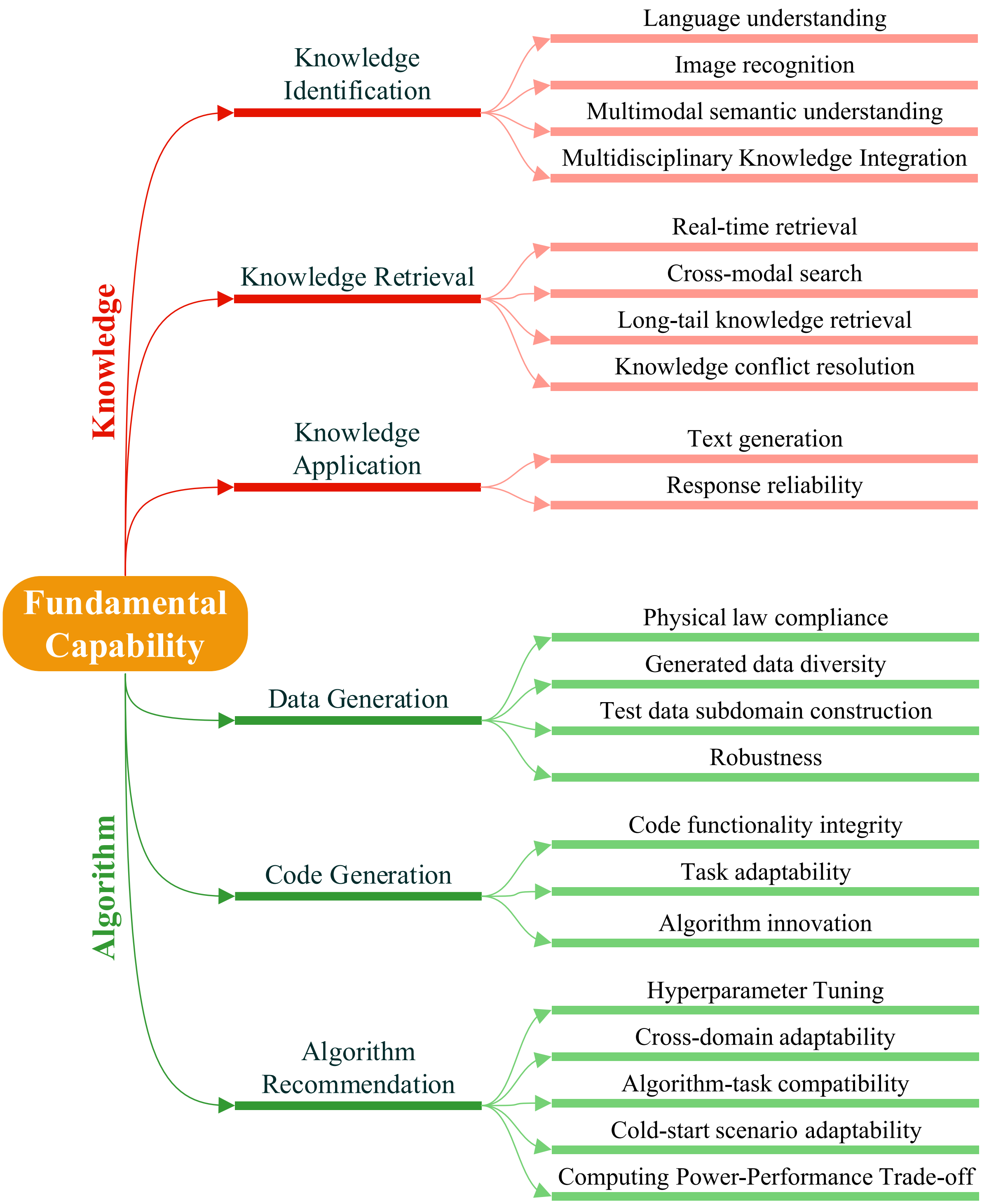

**Fig. 5** Evaluation Metric System for Foundational Capability Dimension

#### 3.2.2.1 Knowledge Sub-Dimension

This sub-dimension is divided into three stages: Knowledge Comprehension, Knowledge Retrieval, and Knowledge Application, encompassing the entire lifecycle of PHM knowledge processing from input to adaptation.

**(1) Knowledge Comprehension**

PHM tasks involve specialized terminology and multimodal data (Kwak & Lee, 2022). This dimension assesses the model's ability to recognize and understand domain-specific terms, resolve semantic confusion, and interpret their contextual meaning. Key evaluation indicators include:

**1) Term Recognition Accuracy (TRA)**

$$TRA = \frac{N_{\text{recognized}}}{N_{\text{terms}}}$$

In which:

$N_{\text{recognized}}$ : The number of PHM terms correctly recognized by the model;

$N_{\text{terms}}$ : Total number of annotated terms in the test set.

To effectively assist PHM system design and development, LLMs must be grounded in comprehensive PHM domain knowledge(L. Yang et al., 2024b) and be capable of accurately interpreting domain-specific and infrequently used terms (e.g., “SOC”, “CBM”)—a prerequisite for solution synthesis, verification, and higher-level knowledge reasoning.

**2) Conflict Detection Accuracy (CDA)**

$$CDA = \frac{N_{\text{detected}}}{N_{\text{conflicts}}}$$

In which:

$N_{\text{detected}}$ : The number of pairs of conflicts successfully identified by the model;

$N_{\text{conflicts}}$ : The total number of real conflict pairs in the test set.

PHM knowledge is often multisource and heterogeneous. Different systems or data processing methods can produce conflicting descriptions (Wu et al., 2023; Xuan Z;Wang, 2024). The model must identify such inconsistencies and support consistency checks and knowledge fusion for downstream diagnosis and decision-making.

**3) Comprehensive Knowledge Capability Index (CKCI)**

$$CKCI = \sum_{i=1}^{n} w_i \cdot A_i$$

In which:

$n$ : The number of sub-disciplines involved;

$A_i$ : The answering accuracy rate of the model in the $i$-th sub-discipline;

$w_i$ : The weight of the $i$-th discipline, adjusted as needed to satisfy, $\sum_{i=1}^{n} w_i = 1$.

Effective PHM requires integrating interdisciplinary and multimodal knowledge. The Comprehensive Knowledge Capability Index (CKCI) evaluates the model's breadth and depth across fields, as well as its potential to evolve with domain knowledge (Bouhadra & Forest, 2024), which is the core of integrated evaluation.

**(2) Knowledge Retrieval**

In the knowledge processing dimension, the knowledge retrieval component is responsible for accurately and efficiently analyzing, retrieving, and ranking user queries. Given the diversity and complexity of PHM knowledge, a comprehensive evaluation of large models must include their performance across standard queries, cross-modal queries, and long-tail knowledge retrieval. For standard queries, conventional metrics such as response time, throughput, accuracy, precision, recall, and F1 score are widely adopted. However, considering the specific demands of PHM, such as cross-modal information handling and the need for generalization, we propose the following domain-specific indicators:

**1) Cross-modal Fusion Efficiency (CFE)**

$$CFE = \frac{P_{\text{task}}}{T_{\text{infer}}} \ (1); \quad CFE = \frac{P_{\text{task}}}{R_{\text{usage}}} \ (2)$$

$$CFE\ Gain = \frac{CFE_{multi\text{-}modal} - CFE_{single\text{-}modal}}{CFE_{single\text{-}modal}}$$

In which:

$P_{\text{task}}$ : Task-related performance indicators, such as fault identification accuracy, F1 score, AUC, RMSE, etc., are set according to specific tasks.

$T_{\text{infer}}$ : The average inference time of the model during testing (unit: seconds).

$R_{\text{usage}}$ : Resource consumption, such as GPU memory usage, FLOPs, energy consumption, etc., is selected according to the actual platform.

Traditional evaluations of cross-modal knowledge fusion primarily focus on retrieval or matching accuracy (Cheng et al., 2022; Xing CL;Lv, 2024), which fail to capture a model's overall efficiency in balancing performance and resource consumption. The Cross-modal Fusion Efficiency (CFE) provides a more comprehensive measure by assessing both task completion quality and resource efficiency during the processing of complex heterogeneous information. Two forms of CFE—performance-to-time efficiency and performance-to-computation efficiency—can be flexibly applied depending on the experimental context, highlighting the mode's practical value in cross-modal retrieval and processing.

**2) Domain-Specific Retrieval Accuracy (DSRA)**

$$DSRA = \frac{N_{\text{correct}}}{N_{\text{queries}}}$$

In which:

$N_{\text{correct}}$ : The number of queries for which the model successfully retrieves the correct knowledge points;

$N_{\text{queries}}$ : The total number of domain retrieval queries initiated.

Unlike general-purpose text search, retrieval tasks in PHM focus on the accurate recall and identification of specialized domain knowledge, such as component details, failure modes, and maintenance strategies. LLMs must go beyond superficial keyword matching and demonstrate the ability to retrieve and understand contextually relevant expert knowledge in response to targeted prompts.

**3) Long-Tail Fault Retrieval Capability (LFRC)**

$$LFRC = \frac{\sum_{i=1}^{n} w_i \cdot r_i}{\sum_{i=1}^{n} w_i}$$

In which:

$r_i$ : the proportion of the long-tail fault class retrieved, which is 0 or 1;

$w_i$ : the importance weight of the $i$-th fault class. The higher the rarity, the greater the weight.

In PHM, fault knowledge is often distributed unevenly—frequent failures versus rare, sudden ones like long-tail faults. Long-tailed fault knowledge is difficult in fault prediction and identification, while LLMs should retrieve effective knowledge of rare faults.

**4) Temporal Knowledge Retrieval Accuracy (TKRA)**

$$TKRA = \frac{N_{\text{aligned}}}{N_{\text{temporal_queries}}}$$

In which:

$N_{\text{aligned}}$ : Number of queries for which knowledge points with matching time windows are successfully retrieved;

$N_{\text{temporal_queries}}$ : Total number of time-sensitive queries.

PHM involves time-dependent data, such as aging curves or health trends (Kumar et al., 2023; J. Wang et al., 2024). Temporal Knowledge Retrieval Accuracy (TKRA) evaluates whether the model can fetch information relevant to specific time windows.

**(3) Knowledge Application**

In the knowledge application sub-dimension, the key capability lies in the generation of text that adheres to PHM-specific standards and conventions. The model should demonstrate the ability to apply retrieved knowledge to produce responses that are not only coherent and accurate but also aligned with industry norms. To evaluate whether LLMs can flexibly apply domain knowledge in reasoning and decision-making, and whether their generated content truly meets PHM requirements, we introduce the following indicator:

**1) Knowledge Consistency Score (KCS)**

$$KCS = \frac{1}{n}\sum_{i=1}^{n}\left(\alpha \times S_{\text{semantic},i} + \beta \times S_{\text{structure},i} + \gamma \times S_{\text{expression},i}\right)$$

In which:

$n$ : Total number of test samples;

$S_{\text{semantic},i}$ : Semantic consistency score of the $i$-th sample;

$S_{\text{structure},i}$ : Structural consistency score of the $i$-th sample;

$S_{\text{expression},i}$ : Normative score of language expression of the $i$-th sample;

$\alpha, \beta, \gamma$ : weights of each sub-index, satisfying $\alpha + \beta + \gamma = 1$, set according to actual

evaluation requirements.

In PHM field, it's not enough to generate grammatically correct text. Outputs must reflect task intent and retrieved knowledge with precision. The Knowledge Consistency Score (KCS) measures how well the model applies retrieved knowledge to generate accurate and coherent responses.

**2) Knowledge Traceability Rate (KTR)**

$$KTR = \frac{N_{\text{traceable}}}{N_{\text{facts}}}$$

In which:

$N_{\text{traceable}}$ : The number of pieces of information in the output that can be accurately traced back to the source of knowledge;

$N_{\text{facts}}$ : The total number of factual pieces of information included in the output.

Considering the issue of model hallucination caused by the training mechanisms of large models (Z. Ji et al., 2023), it is essential to ensure that the knowledge points in the output are real and accurate, with traceable and reliable sources, when performing knowledge application tasks such as PHM knowledge Q&A, writing PHM system design documents, and technical manuals. This helps to prevent the model from fabricating critical facts and ensures the high-quality implementation of PHM tasks. The KTR metric system evaluates whether key knowledge points in the model output can be clearly traced back to the knowledge base/input documents.

**3) Multimodal Knowledge Integration (MKI)**

$$MKI = \frac{N_{\text{modal_aligned}}}{N_{\text{modal_required}}}$$

In which:

$N_{\text{modal_aligned}}$ : The number of key cross-modal pieces of information correctly integrated in the model output;

$N_{\text{modal_required}}$ : The number of key cross-modal pieces of information to be integrated in the input.

PHM system data often comes from multiple modalities, such as sensor data tables, maintenance manual texts, structural diagrams, etc. MKI is used to evaluate whether the model can integrate information from various modalities to produce a unified and correct output.

#### 3.2.2.2 Algorithm Sub-dimension

Regarding the sub-dimension of algorithm, it can be further subdivided into three parts: data generation, code generation, and algorithm recommendation (Behera et al., 2024; Y. Liu et al., 2023), to evaluate the comprehensive capabilities of large models in supporting the design, development, and optimization of PHM algorithm models.

**(1) Data Generation**

To meet the high-quality data requirements of PHM algorithm models, the PHM large model should be capable of performing data generation tasks, where the input consists of equipment information and data feature knowledge elements, and the output is simulated data that mimics the equipment's behavior under predefined conditions and environments (Tao, Li, et al., 2025). To this end, we propose the following metrics to ensure that the model generates scientific, comprehensive, and high-quality simulated data.

**1) Failure Mode Coverage Rate (FMCR)**

$$FMCR = -\sum_{i=1}^{n} p_i \cdot \log\left(p_i\right)$$

In which:

$n$ : The number of types of failure modes/states generated;

$p_i$ : The proportion of the $i-th$ type of failure mode in the generated data;

Fault diagnosis models should cover various typical and edge failure modes (Goswami & Rai, 2023). FMCR is used to measure whether the large model generates sufficiently diverse and representative fault data, reflecting its perceptual ability and generation coverage ability for the input requirements of PHM tasks. This metric essentially represents the information entropy in the generated data, with a higher value indicating a more balanced distribution and broader coverage.

**2) Physical Rule Consistency (PRC)**

$$\text{PRC} = 1 - \frac{1}{m}\sum_{j=1}^{m}\delta_j$$

In which:

$m$ : The number of physical rules;

$\delta_j$ : The proportion of the $i-th$ physical rule being violated. A value of 0 indicates complete compliance, while a value of 1 indicates complete violation.;

The PRC range is [0, 1], with values closer to 1 indicating better compliance with physical laws.

Generated data must not only be structurally correct but also comply with basic physical laws (Kwak & Lee, 2022; H. Li, Zhang, et al., 2024), such as monotonic changes, vibration-temperature correlations, and upper and lower bound constraints. PRC is used to measure the scientific validity of the generated data in terms of physical rationality.

**3) Data Diversity Coefficient (DRC)**

$$DDC = \frac{\sigma_{\text{intra}}}{\sigma_{\text{inter}} + \varepsilon}$$

In which:

$\sigma_{\text{intra}}$ : The average variance within samples of the same type;

$\sigma_{\text{inter}}$ : The average variance between samples of different categories;

$\varepsilon$ : A small constant to prevent division by zero.

The high-quality data generated by the large model for PHM tasks should have sufficient variability to cover its practical needs, such as multi-condition and multi-sample scenarios. DDC is used to evaluate the diversity of the generated data across dimensions such as condition categories and sensor channels.

**4) Subdomain Construction Capability (SDCC)**

$$SDCC = \frac{N_{\text{valid_subdomains}}}{N_{\text{target_subdomains}}}$$

In which：

$N_{\text{valid_subdomains}}$ : The average variance between samples of different categories;

$N_{\text{target_subdomains}}$ : The number of distinct valid subdomains that can be clearly distinguished in the generated data.

In practical engineering, complex PHM testing tasks often need to be decomposed into multiple testing profiles. The large model should be capable of generating the data required for each specific testing profile (Tao, Li, et al., 2025). SDCC measures whether the model can clearly distinguish and generate data for different testing profiles.

**5) Robustness to Perturbation (RBS）**

$$RBS = 1 - \frac{1}{k}\sum_{i=1}^{k} D\left(G_i, G_i^{'}\right)$$

In which:

$k$ : The number of perturbation test samples;

$G_i, G_i^{'}$ : The dataset generated corresponding to the $i-th$ original instruction and the perturbation instruction;

$D(\cdot\ ,\cdot\ )$ : Distribution discrepancy metrics between generated data (e.g., average Euclidean distance, MMD).

A robust PHM model should maintain stable output even with slight changes in instruction phrasing or minor perturbations in parameters (Alomari & Ando, 2024; Khan et al., 2021). RBS measures the stability of the model in generating data under small input variations.

**(2) Code Generation**

Currently, the development of PHM algorithm models faces a series of challenges, such as high demands for specialized knowledge and coding capabilities. Therefore, it is necessary to conduct a comprehensive evaluation of the underlying code generation capabilities of the large model, using the alignment with PHM task requirements as the core metric to assess the quality of the generated code, which reflects the model's code generation ability. As a result, in addition to general metrics like F1 score, we propose the following specialized metrics:

**1) Code Function Coverage (CFC)**

$$CFC = \frac{N_{\text{implemented}}}{N_{\text{total}}}$$

In which:

$N_{\text{implemented}}$ : The number of functional modules successfully generated and validated through testing;

$N_{\text{total}}$ : The total number of functional modules defined based on the task requirement document or business scenario.

PHM algorithms typically consist of multiple functional modules, such as data preprocessing, model training, and metric evaluation. CFC is used to assess whether the model has covered and implemented the specified task functions in the code generation task. A CFC value closer to 1 indicates that the generated code is more comprehensive and practical.

**2) Task Matching Compatibility (TMC)**

$$TMC = \frac{N_{\text{correct_matched}}}{N_{\text{samples}}}$$

In which:

$N_{\text{correct_matched}}$ : The number of samples where the code generated by the model exactly matches the task intent;

$N_{\text{samples}}$ : Total number of samples.

Generated code should not only be executable but also correctly align with the PHM task objectives (Tao et al., 2024). TMC evaluates whether the model, based on its strong semantic understanding, reasoning, and generation abilities, successfully generates 'semantically fitting' code content according to the input task prompt. The matching assessment mainly includes the consistency of key modules' algorithms, primary structural logic, output interfaces, and other elements with the task description.

**3) Parameter Generalization Index (PGI)**

$$PGI = \frac{P_{avg}}{P_{best}}$$

In which:

$P_{avg}$ : The average model performance of the code under all test conditions, such as different fault types, different data sampling rates, etc.;

$P_{\text{best}}$ : The best performance of the code under these conditions.

Given the strong hyperparameter sensitivity of PHM algorithms(Feng et al., 2024; Hu et al., 2022), PGI can be used to evaluate the robustness of the code generated by the large model under different conditions, and whether it can maintain high performance across various task parameter settings. Typical performance metrics include accuracy, RMSE, F1, etc. The closer the PGI is to 1, the better the generalization ability.

**4) Code Robustness Score (CRS)**

$$CRS = \frac{N_{\text{passed_abnormal}}}{N_{\text{test_cases}}}$$

In which:

$N_{\text{passed_abnormal}}$ : The number of test cases that successfully execute even in abnormal scenarios;

$N_{\text{test_cases}}$ : The total number of designed abnormal test cases (e.g., empty CSV, out-of-range data, etc.)

Since PHM systems are typically deployed in complex equipment and involve various types of complex data, models, and related documents, in practical applications, the code often encounters abnormal situations such as missing data, invalid values, missing files and other disturbances(Qin et al., 2023; K. Zhou, Zhong, et al., 2025). CRS is used to evaluate whether the code generated by the large model can be robustly executed.

**(3) Algorithm Recommendation**

PHM includes multiple core functions such as fault diagnosis, fault prediction, and health status monitoring, involving a wide range of algorithms(Y. Cao et al., 2024b; Che et al., 2024; K. Zhang et al., 2024). Additionally, in practical engineering applications, the engineering objects vary, and the working conditions differ significantly. A major challenge faced by the PHM large model is how to select the most appropriate algorithm from a large pool of PHM algorithms based on the specific task requirements of a given object(C. Lu et al., 2025). To address this, we propose the following metrics to assess its algorithm recommendation ability:

**1) Task Adaptation Coverage (TAC)**

$$TAC = \frac{N_{\text{adapted}}}{N_{\text{defined}}}$$

In which:

$N_{\text{defined}}$ : The number of algorithm recommendation objectives defined in the task requirements;

$N_{\text{adapted}}$ : The number of algorithms successfully recommended and reasonably explained.

PHM involves multiple tasks such as condition monitoring, fault diagnosis, remaining useful life prediction, maintenance decision-making, etc. Each task has certain differences in its conventional processing logic, commonly used metrics, and typical algorithms. Based on this, TAC measures whether the large model can match the appropriate algorithm approach or name based on the task description, reflecting its task understanding and algorithm adaptation ability.

**2) Cold-start Scenario Adaptability (CSA)**

$$CDA = \frac{N_{\text{cross_generalized}}}{N_{\text{cross_queries}}}$$

In which:

$N_{\text{cold_matched}}$: The total number of task descriptions that include cold start features;

$N_{\text{cold_cases}}$: The number of algorithms recommended by the model for cold start.

Issues such as limited data and incomplete samples are common while intractable in the PHM field(Lian et al., 2025). The demand for the implementation of typical PHM tasks, such as small-sample fault diagnosis in practical engineering, is growing. Therefore, for algorithm recommendation in large models, the CSA metric can be used to evaluate whether the model can recognize 'cold start' scenarios and, based on the recognition results, recommend algorithms suitable for small-sample/zero-sample conditions (Kojima et al., 2022).

**3) Algorithm-Task Compatibility Score (ATC)**

$$ATC = \frac{1}{n}\sum_{i=1}^{n} Sim\left(T_i, A_i\right)$$

In which:

$n$: Total number of tasks;

$T_i$: Semantic embedding of the $i-th$ task;

$A_i$: Semantic embedding of the recommended algorithm;

$Sim(\cdot)$: The semantic matching score between the two (e.g., using Cosine or BERT embedding similarity).

Given the task-specific characteristics of PHM applications, algorithmic suitability may vary across diagnostic, predictive, and decision-making tasks(T. Li et al., 2022; W. Li et al., 2022). To quantify the appropriateness and adaptability of algorithm recommendations, the ATC score is employed as a unified evaluation metric.

**4) Compute-Predict Tradeoff Score (CPT)**

$$CPT = \frac{S_{perf}}{S_{cost} + \varepsilon}$$

In which:

$S_{perf}$ : Performance metrics of the recommended algorithm for the same task (e.g., F1, RMSE);

$S_{cost}$ : Inference resource cost (e.g., number of parameters, computation time);

$\varepsilon$ : Small value to prevent division by zero.

In practical engineering, the design and implementation of PHM models require a comprehensive balance between accuracy and computational resource constraints. CPT is used to measure the overall score of the recommended algorithm in the 'resource-performance' dimension.

### 3.2.3 Entire lifecycle dimension

As the core component of the PHM system, the PHM model is fundamental to realizing health management throughout the equipment's entire lifecycle. Therefore, the performance of the PHM large model should serve the overall goal of health management across the entire lifecycle of the equipment. The evaluation system for the PHM large model must comprehensively consider the PHM requirements across the entire lifecycle.

From the perspective of the current PHM system's entire lifecycle validation framework, existing studies often employ a combination of the four characteristics—reliability, maintainability, supportability, and testability—as the validation method. Based on this, the validation content is divided into two major parts: system component validation and system algorithm validation. This approach has been widely applied in fields such as aerospace, shipbuilding, and military equipment, generating a wealth of engineering experience and related metrics. However, due to the wide coverage of PHM systems, the diversity of tasks, and the dynamic complexity of operating conditions, these validation metrics are currently fragmented, modularized, and lack a unified task-oriented structure. This makes it difficult to support cross-phase, integrated systemic capability analysis. Particularly in the context of the growing demand for PHM capabilities, there is an urgent need to establish a systematic organizational structure for metrics, with tasks as the main thread, phases as the foundation, and metrics as the support, to drive the transition from traditional validation metrics to a capability evaluation dimension framework. Therefore, this paper proposes the development of a large model evaluation focusing on the entire lifecycle dimension of PHM.

The entire lifecycle dimension is essentially not an entirely independent capability evaluation dimension, but serves as the logical main thread and goal orientation of the PHM-Bench evaluation system design. It systematically integrates the aforementioned “fundamental capability dimension” and “core task dimension” for capability decomposition and metric design. Therefore, this section systematically organizes the assessment requirements and the distribution of the metric system for each stage of the equipment's entire lifecycle based on the basic work items of PHM business realization, existing lifecycle validation metrics, and methods. It also reconstructs the organization based on key characteristics such as task coverage, phase adaptability, and metric standardization to reflect the entire lifecycle engineering alignment of the metrics proposed in this paper, as shown in Fig. 6.

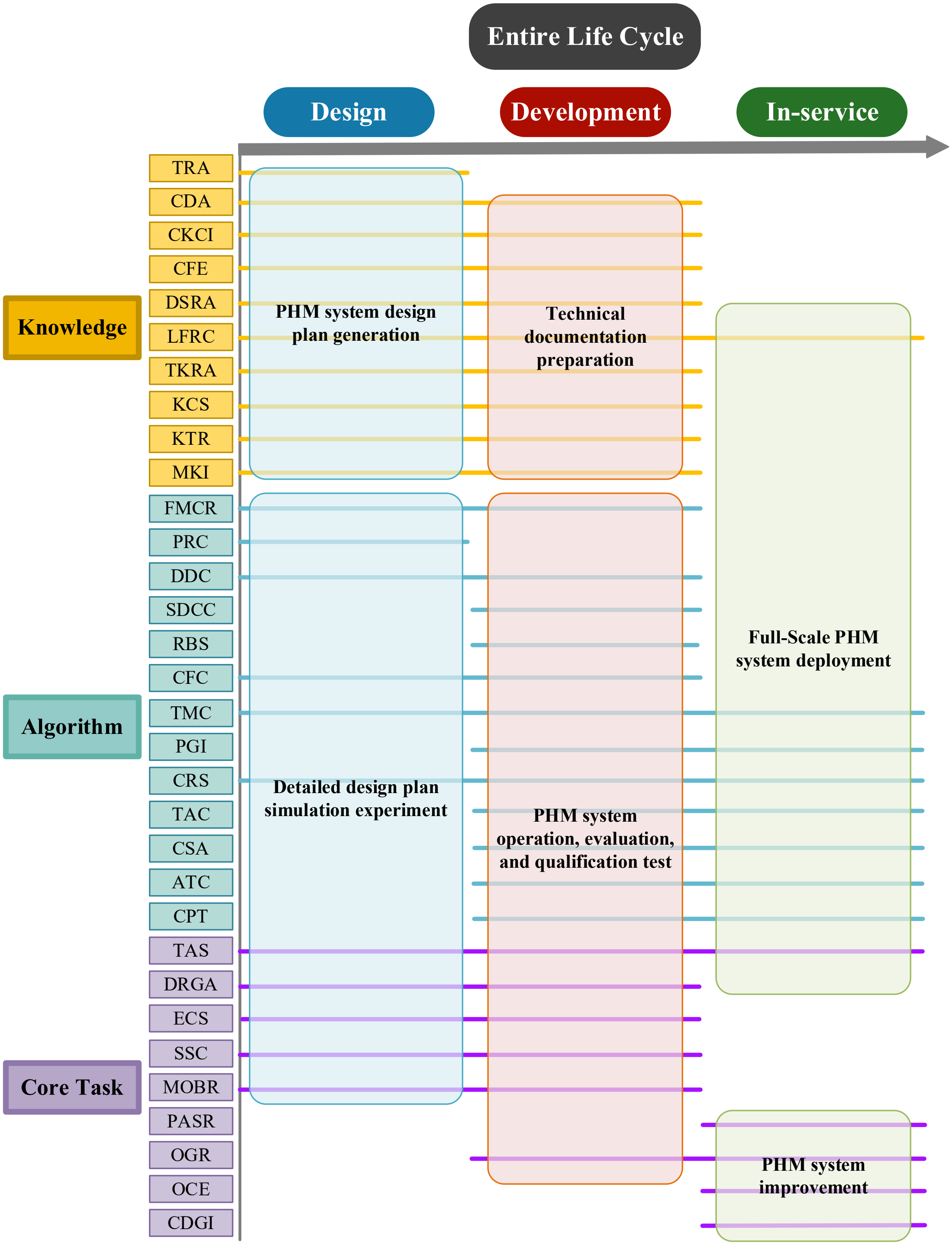

**Fig. 6** Entire lifecycle Dimension Evaluation Metric System

Fig. 6 presents the “Phase Adaptation - Typical Tasks - Verification Metrics” mapping system built by PHM-Bench across the entire lifecycle dimension. This system aims to systematically and interpretatively quantify the performance of PHM large models in terms of their capabilities to support tasks in different stages of the lifecycle. The system is structured around three core elements: verification metrics (vertical), lifecycle stages (horizontal), and task examples (embedded). It comprehensively covers all stages of the equipment lifecycle, including design, development, and operation. The system also systematically considers the stage applicability of the dimensional classifications and metrics system introduced in the first two sections of this paper. In addition, it distributes the tasks based on the evaluation needs of scenarios such as equipment design analysis, experimental validation, and operational use, in order to objectively and effectively reflect the consistency between model capabilities and engineering verification systems.

## 3.3 Evaluation Method and Assessment Process

To verify the rationality of the aforementioned evaluation framework and the effectiveness of the evaluation metrics, and to ensure the universality, flexibility, and systematization of the evaluation in practical industrial scenarios, it is recommended to adopt a combined evaluation approach of automated assessment and expert evaluation. The automated assessment mainly relies on fine-tuned super-large models and the PHM domain-specific metrics provided in Section 2 to quantitatively score the model outputs. Expert evaluation involves inviting domain experts to qualitatively and quantitatively review the results of key tasks based on established scoring criteria. This section will introduce both methods one by one and demonstrate the large model PHM capability assessment process based on the "PHM-Bench" evaluation framework. The evaluation is centered around the three-dimensional quantitative evaluation metrics system of PHM-Bench, supplemented by the structured reference framework based on the LLM-as-a-judge evaluation mechanism (G. H. Chen et al., 2024), to cover different PHM task goals and application scenarios. At the same time, it supports user-configurable evaluation dimensions and metrics to achieve flexible and scalable evaluation services.

### 3.3.1 Automated Assessment

The automated assessment module aims to utilize a super-large model scoring system by constructing specialized prompt templates and dedicated evaluation metrics to conduct bulk, standardized, and quantitative reviews of the maintenance plans generated by the model and the results of various sub-tasks. The specific process mainly includes the following four stages:

**(1) Selection of Evaluation Dimensions**

Industrial users select the required evaluation dimensions from the foundamental capability dimension, core task dimension, entire lifecycle dimension, and their sub-dimensions based on the actual PHM task scenarios and model application objectives. Since each dimension contains multiple refined sub-metrics, users can tailor or weight these dimensions based on their needs.

**(2) Evaluation Data Preparation**

It supports the automatic calling of predefined PHM case sets, question sets, and other benchmark datasets and task templates, as well as typical operating condition data, fault sample libraries, task specification documents, and other resources. Using prompt engineering, the evaluation samples are automatically constructed to enable comprehensive testing.

**(3) Automated Execution of Evaluation Tasks**

The unified evaluation engine is called to automate the entire process of model input, response generation, and evaluation metric calculation output, based on the selected evaluation dimensions and their quantitative metrics. The process supports multiple rounds of interaction, cross-modal data, complex task chains, and other types of tasks, ensuring the systematization and comprehensiveness of the evaluation.

**(4) Evaluation Result Generation**

Based on the calculated dimension matrix $R_{i,j}$, the weighted summation is performed to generate the automated evaluation score:

$$\text{Score}_{\text{auto}} = \frac{1}{N}\sum_{i=1}\left(\sum_{i=1} W_j \cdot R_{i,j}\right)$$

In which:

$i$ : Task index;

$j$ : Evaluation dimension;

$W_j$ : Weight of each evaluation dimension, adjusted as needed;

$N$ : Sample size.

At the same time, multidimensional evaluation results and their visual displays are automatically generated, supporting personalized metric ranking, historical comparison analysis, performance radar charts, and other forms. This allows users to perform quantifiable diagnostics and iterative optimization of the model's performance.

### 3.3.2 Expert Evaluation

To ensure the objectivity and interpretability of the automated evaluation, the expert evaluation module invites multiple senior experts and practitioners from the PHM field to review several core competencies that require qualitative judgment. We refer to the LalaEval (Sun et al., 2024) review method and design the specific process as follows:

**(1) Formation of Expert Panel**

Several domain experts are invited to review the automated evaluation results and provide detailed feedback on certain task samples, evaluation dimensions, or evaluation capabilities that require qualitative judgment or are controversial after quantitative judgment.

**(2) Evaluation Dimensions and Metrics**

Based on the selected objects and feedback results, evaluation criteria are set. Standard Likert scales or a 0-3 scoring system may be used, as shown in Table 1.

**Table 1:** Expert Evaluation Mechanism

| Evaluation Item | Scoring Mechanism | Evaluation Criteria |
|---|---|---|
| Physical Law Conformance | 0-3 | Evaluates whether it conforms to basic scientific or empirical laws |
| Code Interpretability | 0-3 | Assesses whether the code logic and module functionality are clearly understood |
| Solution Interpretability | 0-3 | Assesses whether the generation basis and structure of the solution are clearly defined |
| Multidisciplinary Integration Ability | 0-3 | Evaluates whether reasonable cross-domain knowledge integration is possible (e.g., mechanical + materials) |
| Algorithm Innovation | 0-3 | Assesses whether the method has research or engineering innovation value |
| Task Adaptability | 0-3 | Evaluates whether it can accurately match the PHM subtask intentions and background |
| Generalization Ability | 0-3 | Assesses the output rationality in cross-task/cross-object scenarios |

**(3) Evaluation Data Collection**

We take single blind test to ensure the fairness of the evaluation. Experts provide scores based on evaluation criteria (which are essentially the same as the automated evaluation dimensions). For expert ratings, we refer to the Borda Count method (de Pater & Mitici, 2023; Jose, Nguyen, Mediaher, Zemouri, Lévesque, et al., 2024; Kamariotis et al., 2024; Kumar et al., 2023; Su et al., 2024; Yan et al., 2021b; Yang et al., 2025) to compute the scores as follows:

$$Score_{(qj)} = \frac{\sum_{j=1}^{N}\left(\sum_{i=1}^{N} W_j \cdot R_{i,j}\right)}{N}$$

In which:

$q$ : Evaluated model;

$j$ : Evaluation dimension/sub-dimension/ability;

If any evaluator gives a perfect score on all topics for any task, then $Score_{(qj)} = 1$.

Thus, the score $Score_{(qj)}$ is normalized to the interval $[0,1]$, and the model score $Score_{expert}$ is the average of all $Score_{(qj)}$.

### 3.3.3 Evaluation Integration and Result Generation

We integrate the automated evaluation scores with the expert evaluation scores. Based on the foundation of quantitative indicators, expert feedback is used to further refine and adjust the scores, ensuring consistency between the automated evaluation and expert scores. The final composite score is calculated as:

$$\text{final Score} = \alpha \cdot \text{Score}_{\text{auto}} + \beta \cdot \text{Score}_{\text{expert}}, \quad \alpha + \beta = 1$$

In which:

$\alpha$ and $\beta$ represent the weights of the automated and expert evaluations, respectively. It is suggested that the initial values be set as $\alpha = 0.7$ and $\beta = 0.3$, with adjustments made based on experimental feedback and task-specific needs.

**(1) Result Presentation**

Based on the evaluation task requirements, the final evaluation report can present the performance of different models in areas such as basic capabilities, core tasks, and entire lifecycle through visual forms like radar charts, bar charts, etc. The report will also list the specific scores for each dimension, expert feedback, and improvement suggestions.

**(2) Model Comparison and Optimization Suggestions**

Using this evaluation process, the advantages and disadvantages of different PHM large model solutions in multi-task coordination and cross-stage capabilities can be compared. This provides a basis for future model optimization and application promotion.

# 4 PHM Evaluation Dataset and Benchmark Scheme Construction

## 4.1 Dataset Design and Construction Framework

### 4.1.1 Framework Overview: Task-Driven Dataset Construction Logic

The PHM evaluation dataset described in this section is essentially a standard task case library corresponding to the task-driven evaluation method outlined in Section 3.3. To support the comprehensive capability assessment of large models in tasks such as fault diagnosis, lifetime prediction, solution generation, and maintenance recommendations within the PHM-Bench evaluation system, all data samples are uniformly encapsulated in a structured JSON format. This ensures consistency of input data, precise matching of capability metrics, and automated execution of the evaluation process.

### 4.1.2 Evaluation Task Composition and Execution Method

The evaluation tasks designed in this study depart from traditional question-answer or single-choice paradigms and instead are driven by actual industrial business scenarios. A structured task chain is constructed as the evaluation input unit. Each task sample includes key elements such as prompts, object information, operating conditions description, data fragments (e.g., time-series signals, images, text), and target requirements. The model's response must cover natural language explanations, mathematical expressions, reasoning logic, and deployable code (for code generation tasks). The evaluation method combines automated performance metrics (e.g., RMSE, CFC, FMCR), large model scoring mechanisms (LLM-as-a-judge), and expert review results. Ultimately, this produces a multidimensional scoring table, radar chart, and performance comparison report that fully reveals the PHM capability performance.

### 4.1.3 Data Source and Composition Structure

To ensure that the dataset has sufficient representativeness, operational complexity, and application breadth, the PHM-Bench task case set mainly consists of the following two parts:

**(1) Structured Task Case Set:**

**Table 2:** Original Data Sources for the Case Set

| Data Source | Category | Quantity |
|---|---|---|
| High-quality academic papers in PHM | See Table 3, "Object Type" | 16472 (Selected 108 highly relevant papers from the last three years) |
| Fault diagnosis textbooks | \ | 9 |
| PHM field patents | Bearings | 429 |
| | Gears | 230 |
| | Gearboxes | 48 |

**Table 3:** Classification of Case Set Object Types and the Number of Designed Cases

| Object Type | Number of Cases |
|---|---|
| Bearings | 26 |
| Gears | 5 |
| Belts | 3 |
| Structural Components | 3 |
| Propellers | 8 |
| Engines | 10 |
| Electronic Components | 3 |
| Converters | 6 |
| Motors | 3 |
| Communication Systems | 3 |
| Navigation Systems | 2 |
| Transmission Systems | 3 |
| Radars | 5 |
| Tracks | 3 |
| Circuits | 5 |

| Object Type | Number of Cases |
|---|---|
| Batteries | 12 |
| Gases | 3 |
| Other Systems | 7 |

In order to support the unified invocation of the PHM-Bench evaluation system, model response parsing and multi-dimensional index evaluation, we designed a multi-dimensional labeling system, defining standardized labeling fields for all task samples, such as "Case background. Component or system description. Specific object", "Work condition description. Normal operating conditions. Characteristics", "Data Description. Data source. Collected data. Data characteristics. Sampling frequency", etc. The labeling system involves 7 dimensions and 18 levels, such as task objective, data description, working condition analysis, data processing analysis, model evaluation analysis, etc., to generate a test case set that systematically covers the core tasks, key issues, and requirements of PHM.

**(2) Open source industrial data subsets:**

We integrated and extended typical datasets such as XJTU-SY, FEMTO-ST, PU, Paderborn, etc., covering the typical operation and fault states of bearings, motors, and other key equipment, as shown in Table 4. The raw data are transformed into input units that meet the PHM-Bench measurement task criteria through scene reconstruction and semantic tag complementation.

**Table 4:** Collation of Open Source Industrial Data Subsets

| **Data set** | **Source Organization** | **Objects** | **Adaptation Tasks** | **Adaptation Instructions** |
|---|---|---|---|---|
| UC gearbox (P. Cao et al., 2018) | University of Connecticut | Pinion (gear wheel) | Diagnostic | Multi-level classification of complex gear failures |
| JULY Motors (Xuelangyun, 2022) | Snow Wave Industrial Challenge | Electrical machinery | Diagnostic | Batch Classification and Comparison |
| DC bearings (industrial-bigdata, | Mathematical Innovation | Bearings | Diagnostic | Diagnosis of complex state combinations |

| Data set | Source Organization | Objects | Adaptation Tasks | Adaptation Instructions |
|---|---|---|---|---|
| 2024) | Competition | | | |
| IMS bearings (Qiu et al., 2006) | University of Cincinnati IMS Center | Bearings | Condition monitoring/ Prediction | Life Prediction and Degradation Modeling |
| CWRU Bearing (Case Western Reserve University, n.d.) | Caesarea University | Bearings | Diagnostic | Classification of fault types and severity |
| PU bearings (Lessmeier C, 2016) | University of Paderborn | Bearings | Diagnostic/ Condition monitoring | Multimodal fusion study |
| UO bearings (Huang & Baddour, 2018) | University of Ottawa | Bearings | Diagnostic | Dynamic condition and timing modeling |
| JNU bearings (K. Li et al., 2017) | Jiangnan University (Jiangsu Province) | Bearings | Diagnostic | RPM Sensitivity Analysis |
| MFPT bearings (Society for Machinery Failure Prevention Technology, n.d.) | MFPT Association | Bearings | Diagnostic | Feature learning under load variation |
| RIO Bearing (Federal University of Rio de Janeiro, n.d.) | Federal University of Rio de Janeiro | Bearings | Diagnostic | Large-scale generalization tests |
| XJTU bearing (B. Wang et al., 2020) | Western University of Communications | Bearings | Diagnostic/ Predictive | Prediction and Fault Modeling Validation |
| NASA electric actuator (NASA, n.d.) | NASA | Electric actuator | Diagnostic/ Predictive/ | Multi-Parameter Fusion and Complex |

| Data set | Source Organization | Objects | Adaptation Tasks | Adaptation Instructions |
|---|---|---|---|---|
| | | | Condition monitoring | Fault Modeling |
| AIR compressor (Verma et al., 2016) | IIT Indian Institute of Technology | Compactors | Diagnostic | Acoustic Fault Detection Tasks |
| IIT drill (D. Ji et al., 2021) | IIT Indian Institute of Technology | Driller | Diagnostic | Process monitoring |
| ZJU AUV robot (Koopman & Wagner, 2017) | Zhejiang University | Underwater Micro robot | Diagnostics/ Maintenance Decision-making | System level PHM with state push Reasoning |

To achieve the multi-source data adaptability of the PHM-Bench evaluation framework, we carry out systematic label structure extraction and standardization conversion for the above typical open-source industrial data subsets. Unlike the structured task case set, which is designed to uniformly construct label fields at the beginning, for the existing open-source data subset, we uniformly align the PHM-Bench evaluation input specifications through the steps of label field mapping, task adaptation conversion, work condition and original information completion, and automatic template encapsulation.

## 4.2 Measuring Experimental Benchmark Design and Implementation

In order to systematically evaluate the multi-dimensional capability performance of the grand model in PHM tasks, PHM-Bench constructs an experimental baseline based on a three-dimensional measurement system (fundamental capability dimension, core task dimension, and Entire lifecycle dimension). Each dimension baseline corresponds to a representative SOTA grand model and an industrial general model in the current domain, and evaluates their applicability and performance on different tasks. For example, GPT-4o (Shahriar et al., 2024)for natural language processing, Gemini 1.5 Pro (Sonoda et al., 2024)for multimodal processing, Qwen2.5-72B-instruct (Imtiaz Ahmed, 2025) forcustomized deployment, and so on, which is used as a cross-comparison reference to support large model evaluation and performance diagnosis under multiple metrics, multiple tasks, and multiple objects.

**Table 5:** Baseline of Fundamental Capacity Dimension

| Subcapacity | Baseline model | Competency performance reference |
|---|---|---|
| Semantic cognition | GPT-4o<br>ChatGLM3<br>Qwen2.5-72B | TRA=0.87<br>TCCR = 0.82<br>CDA=0.87<br>CKCI=0.84 |
| Cross-modal information processing | Gemini 1.5 Pro<br>-Qwen-VL-32B-Instruct | CFE= 0.76 (Task F1:0.91, Delay:1.2s)<br>MKI=0.79 |
| Knowledge retrieval | GPT-4o<br>Qwen2.5-72B<br>ChatGLM3-6B-Finetune (PHM) | DSRA = 0.88<br>LFRC=0.76<br>TKRA=0.83 |
| Report/Document Generation | GPT-4o<br>Gemini 1.5 | KCS = 0.91<br>KTR=0.88 |
| Code Generation and Parameter Generalization | Code Llama<br>PanGu-Coder2<br>Claude 3 Opus | CFC=0.84<br>TMC=0.82<br>PGI=0.65<br>CRS = 0.66 |
| Algorithm Recommendations | GPT-4o + Toolformer fine tuning<br>Gemini 1.5 Pro<br>Claude 3 Opus | TAC = 0.78<br>CSA=0.65<br>ATC = 0.83<br>CPT = 0.73 |
| Data generation | Qwen2.5-72B-instruct<br>GPT-4o + fine-tuning | FMCR=0.80<br>PRC=0.77<br>DDC = 0.70<br>SDCC = 0.72<br>RBS = 0.71 |

**(2) Core Task Dimension**

The core task dimension focuses on the model's ability to perform professionally in key PHM tasks such as fault diagnosis, life prediction, solution recommendation, and decision support. The reference benchmarks are listed below:

**Table 6:** Baseline of Core Task Dimension

| **Subtask** | **Baseline model** | **Competency performance reference** |
|---|---|---|
| Fault diagnosis | GPT-4o<br>Claude 3 Opus | Accuracy= 0.88, F1= 0.85<br>TAS = 0.82<br>DRGA = 0.85<br>ECS = 0.76<br>SSC=0.83 |
| Fault & RUL prediction | DeepRUL<br>LSTM<br>ChatGLM | RMSE≈320~550 cycles<br>SSC=0.81 |
| Maintenance scheme generation | GPT-4o<br>Claude 3 Opus | MOBR=0.73 |
| Multi-option selection | Generic LLM + RAG | SSC=0.81 |
| Program Optimization | Gemini 1.5 Pro<br>GPT-4o<br>Claude 3 Opus | PASR = 0.79<br>OGR=0.72<br>OCE = 2.3 rounds/task<br>CDGI=0.64 |

The above models represent the three mainstream paradigms of traditional industrial algorithms, deep learning methods, and large language models for PHM tasks, and can be used to evaluate the adaptability and performance differences of different types of models under the three-dimensional structure of “complex task - multiple inputs - highly constrained outputs”.

**(3) Entire Lifecycle Dimension**

The entire lifecycle dimension emphasizes the task generalization and continuous iteration capability of the large model in the whole process of design, development, and service. On the one hand, its capability requirements and quantitative logic have been systematically embedded in the capability decomposition and index design of the aforementioned "fundamental capability dimension" and "core task dimension", such as the large model's capability of understanding, generating and optimizing different tasks is reflected in its adaptability to the tasks in each phase of the whole life cycle of the PHM system, and the large model's ability of responding to the input types, changes in the working conditions, and migration of the objects reflects its responsiveness to the application scenarios in the service phase of the PHM system. On the other hand, as a PHM system-level evaluation dimension, there are currently no unified testing conditions for all phases and tasks, and the combination of verification methods and related indexes only for the "four properties" of the PHM system or for a single component-level object have been widely used in the fields of aerospace, marine, and weaponry equipment, etc., and supported by a large amount of engineering practical experience. In view of this, this study will not set up a separate test mechanism for this dimension.

### 4.2.2 Measurement Experiment Implementation Process

**(1) Fundamental Capability Dimension Assessment Methods and Processes**

The fundamental capability dimension assessment takes the arithmetic sub-dimension as an example, and is mainly oriented to the application scenarios involving algorithm development, code generation, and model optimization in the PHM domain, such as industrial algorithm research and development, model performance evaluation, and validation of cross-task migration capability. The evaluation process of this dimension includes four phases: input and code generation, code verification and execution, code quality assessment, and generalization and robustness testing, which systematically evaluates the engineering capability of the large model in algorithm development in the PHM domain, and the specific process is shown in Fig. 7.

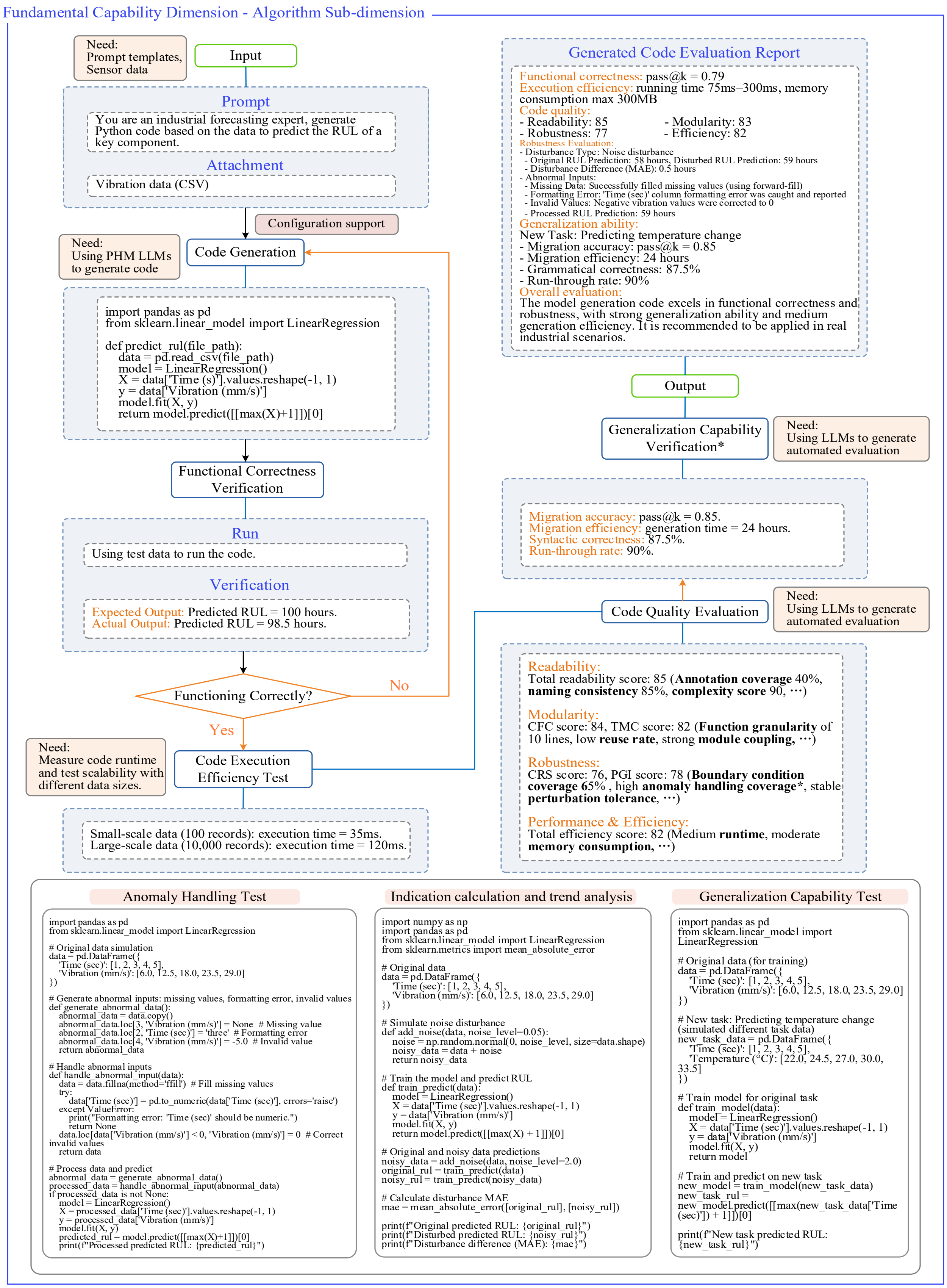

**Fig. 7** Example of PHM-Bench Fundamental Capability Dimension Measurement

This process enables PHM-Bench to systematically assess the quality, usability, generalizability, and robustness of PHM-related code generated by large models, thereby offering a reliable basis for model evaluation and facilitating their integration into practical industrial workflows.

**(2) Core Task Dimension Measurement Methods and Processes**

The core task dimension focuses on the demand for highly complex, multi-constraint tasks in industrial scenarios and evaluates the model's capability in the tasks of plan generation, plan selection, and plan iteration optimization. The evaluation process of this dimension mainly includes four parts: input and task definition, plan generation and evaluation, optimization iteration, and performance evaluation, aiming to systematically verify the effectiveness and engineering applicability of the model in complex decision support.

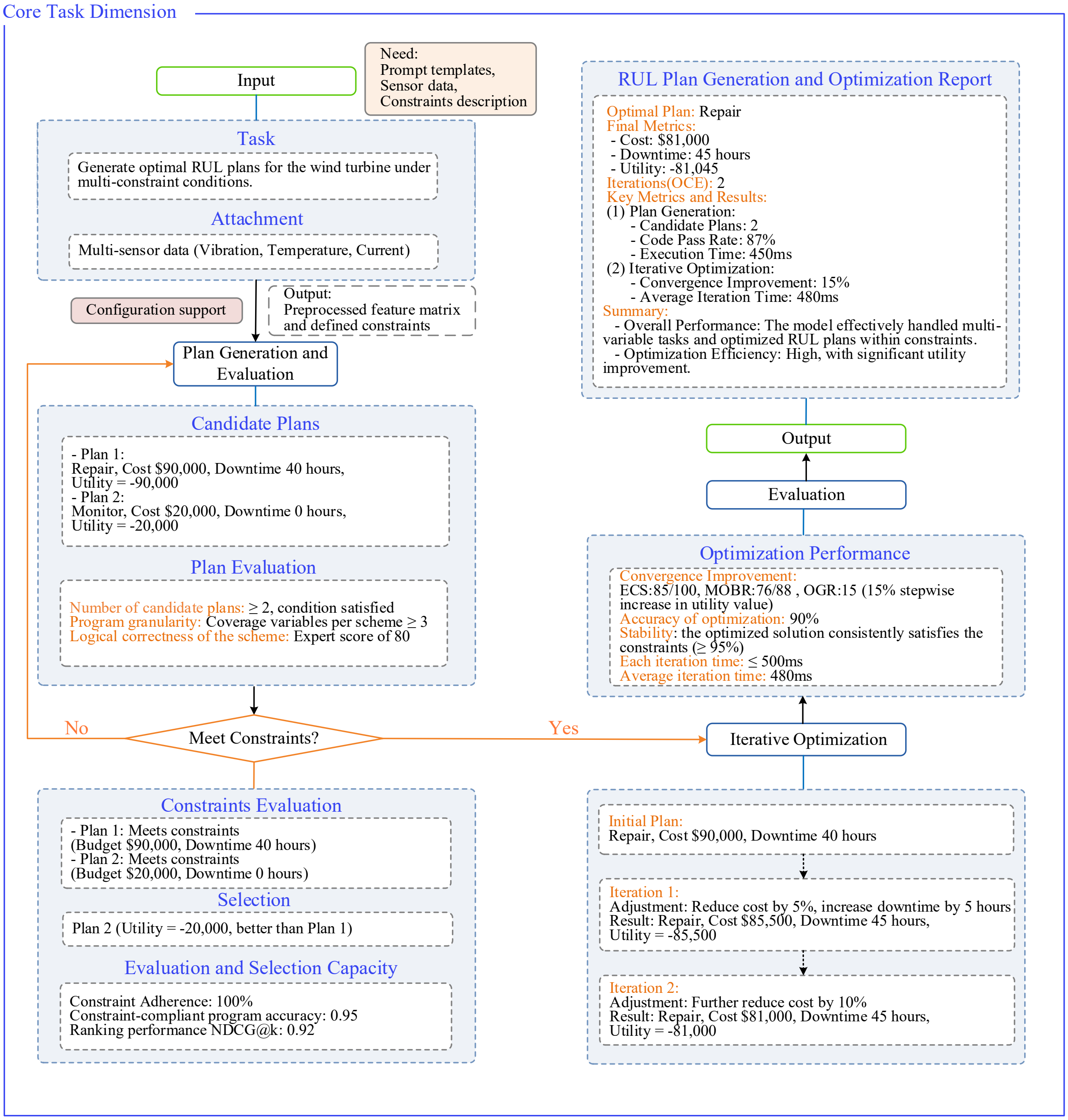

**Fig. 8** Example of PHM-Bench core task dimension measurement process

The above process demonstrates a comprehensive performance evaluation method for the generation, selection, and iterative optimization capabilities of PHM task scenarios for large models under constraints, and the optimization capabilities and execution efficiency of the models in complex task scenarios are comprehensively verified through the core task dimension measurements.

These assessment cases support the validity, interpretability, and domain alignment of the PHM-Bench evaluation framework, suggesting its potential for robust and consistent benchmarking in PHM contexts.

## 5. Conclusion

This study addresses the systematic assessment needs of large models in the PHM domain and proposes PHM-Bench, a domain-oriented evaluation framework with theoretical grounding and practical adaptability. As the first comprehensive assessment framework and engineering practice in the PHM field, PHM-Bench establishes a quantifiable and extensible evaluation system, incorporating standardized datasets, automated assessment pipelines, and an expert review mechanism. This framework helps fill the existing gap caused by the lack of unified standards and systematic evaluation methodologies for large models in the PHM field. Through a series of empirical experiments and baseline construction, the proposed framework has demonstrated its effectiveness in model comparison, capability diagnosis, and optimization guidance. PHM-Bench contributes both theoretical insights and practical tools for building standardized, interpretable, and reproducible assessment systems, which will play an important role in promoting the development of large models in the field of PHM.

Future work should focus on refining the indicator system, diversifying test scenarios, enhancing automation across the evaluation workflow, and strengthening the correlation between assessment outcomes and real-world engineering value. These efforts will provide stronger support for the intelligent health management of high-reliability and high-complexity industrial systems.